\documentclass[letterpaper]{article} 
\usepackage{aaai2027}  
\usepackage[hyphens]{url}  
\usepackage{graphicx} 
\usepackage{natbib}  
\usepackage{caption} 
\usepackage{algorithm}
\usepackage{algorithmic}
\usepackage{amsmath}
\usepackage{enumitem}
\usepackage{amsmath}
\usepackage{amsthm}
\usepackage{amssymb}
\usepackage{lineno}

\newcommand{\up}{\ensuremath{\uparrow}}
\newcommand{\down}{\ensuremath{\downarrow}}

\usepackage{newfloat}
\usepackage{listings}

\usepackage{xspace}
\newcommand{\methodname}{\textbf{SAIL}\xspace}
\DeclareCaptionStyle{ruled}{labelfont=normalfont,labelsep=colon,strut=off} 
\floatstyle{ruled}
\newfloat{listing}{tb}{lst}{}
\floatname{listing}{Listing}

\usepackage{booktabs}

\title{Disentangled Skill Representations for Predictive Human Modeling}
\author{
    Mariah Schrum,
    Deepak Gopinath,
    Srijan Srivatsa,
    Guy Rosman,
    Tiffany Chen
}

\affiliations{
    Toyota Research Institute\\
    Los Altos, California, USA
}

\begin{document}

\maketitle

\begin{abstract}
Understanding human skill is important for AI systems that collaborate with, coach, or assist people. Unlike typical latent variable estimation problems which rely on single observations, skill is a persistent, compositional, and behaviorally grounded construct that must be inferred from patterns over time. We introduce Skill Abstraction with Interpretable Latents (\methodname), a method for modeling human skill as an interpretable, multi-dimensional construct inferred from naturalistic behavior. Our approach produces a skill embedding that is robust to transient performance fluctuations and learns a transferable representation of human subskills. Furthermore, \methodname supports skill-informed behavior prediction that generalizes across a variety of in-domain contexts. We represent each individual with a persistent skill embedding that controls a blend between expert and novice  bases and is trained using counterfactual subskill swaps for disentanglement. This design encourages representations that are both robust to performance variation and structured for interpretability. We demonstrate across racing and baseball that \methodname{} achieves
strong predictive performance and consistently improves
behaviorally grounded disentanglement over the evaluated baselines,
while also improving downstream AI coaching performance.
\end{abstract}


\section{Introduction}
\label{sec:intro}

AI systems that support, collaborate with, or coach humans must reason about human skill to personalize instruction, anticipate behavior, and adapt assistance over time. Unlike many latent variables in machine learning, human skill cannot be inferred from individual actions or outcomes. Instead, it is a persistent, behaviorally grounded, and compositional construct that must be inferred from patterns across repeated interactions while accounting for noise, variability, and changing task conditions \citep{iso2024theory,langley2004hierarchicalskills}.

Human skill differs from many notions of ``skill'' used in machine learning. In robotics and reinforcement learning, skill often refers to reusable action primitives or policies for task execution \citep{lesort2018state}. In contrast, we model human skill as a persistent, participant-level construct composed of multiple interpretable subskills \citep{newell1991motor,ericsson1993role}. We further distinguish skill from performance: performance reflects trial-specific outcomes influenced by situational factors such as fatigue or risk-taking, whereas skill represents stable abilities that generalize across contexts \citep{iso2024theory,fitts1967human}. Conflating the two can lead to inaccurate assessment and inappropriate interventions.

We propose that an effective skill representation should satisfy three desiderata:

(1) \textbf{Construct Validity} \citep{messick1995validity}: representations should remain stable across sessions and robust to trial-level noise.

(2) \textbf{Predictive Utility}: representations should support accurate behavior prediction across in-domain contexts.

(3) \textbf{Interpretability}: representations should decompose into disentangled subskills that correspond to human-recognizable aspects of expertise.

To satisfy these desiderata, we introduce \textbf{S}kill \textbf{A}bstraction with \textbf{I}nterpretable \textbf{L}atents (\methodname), a computational framework for representing human skill. Rather than learning an unconstrained latent embedding and hoping it reflects skill, SAIL incorporates inductive biases inspired by theories of skill acquisition: skill is persistent across observations, expressed through behavior, progresses relative to expertise, and is composed of interpretable subskills.


Intuitively, we ask how an individual's behavior differs from characteristic novice and expert behaviors, rather than asking the representation to explain every observed trajectory directly.
This constrains the representation to encode structured, skill-relevant variation instead of  behavioral fluctuations. To encourage interpretability, we supervise subskill-specific latent slices with behaviorally grounded metrics and introduce a counterfactual training procedure that encourages each latent slice to represent a distinct subskill. Importantly, behavior prediction serves as a supervision signal for learning the representation rather than the primary objective.

In this work we contribute the following:

\begin{enumerate}[nosep,leftmargin=*]
\item We formulate human skill modeling as a representation learning problem and identify construct validity, predictive utility, and interpretability as key desiderata.
\item We propose \methodname, a participant-level skill representation that combines participant embeddings, novice--expert basis blending, and counterfactual supervision to learn stable, predictive, and interpretable skill representations.
\item We demonstrate across racing and baseball that \methodname{} outperforms strong baselines and improves downstream instructor-feedback prediction by 10\%.
\end{enumerate}
\section{Related Work}

Human skill has been studied across education, sports science, robotics, and human--AI interaction~\citep{anderson2014learning,ericsson1993role}. Unlike task performance, skill is a persistent, latent construct that must be inferred from behavior accumulated over time rather than individual outcomes~\citep{newell1991motor,schmidt2018motor}. Traditional measures such as completion time or accuracy~\citep{fitts1967human} are highly context dependent and often reflect transient \emph{performance} rather than underlying skill. Psychometric approaches, including Item Response Theory and Bayesian Knowledge Tracing, estimate related latent constructs~\citep{embretson2013item,corbett1994knowledge,piech2015deep}, but are designed for discrete responses rather than continuous behavioral trajectories.

Trajectory-based approaches, including clustering and inverse reinforcement learning, infer latent  structure from demonstrations~\citep{ziebart2008maximum,abbeel2004apprenticeship}. Likewise, work in robot teaching and reinforcement learning often represents "skills" as reusable action primitives or control policies~\citep{argall2009survey,cakmak2012designing,hausman2018embedding,petangoda2019disentangled,dave2025skill}. While effective for policy learning, these methods do not model  skill as a persistent, interpretable construct that generalizes across repeated observations.

Representation learning methods, including autoencoders, variational autoencoders, and contrastive learning, have been widely used to encode human behavior~\citep{kingma2013auto,oord2018representation,zhang2019self}. More recently, participant-level representations have been explored to model persistent latent characteristics across repeated interactions, supporting personalization, adaptive human--AI interaction, and human behavior modeling~\citep{jacques2019social,gopinath2017human,decastro2024dream2assist,jeon2020shared,schrum2023reciprocal_mind_meld}. Disentangled representation learning further seeks to recover interpretable latent factors~\citep{higgins2017beta,chen2016infogan,kim2018disentangling}, although purely unsupervised objectives do not guarantee semantic alignment or identifiability~\citep{locatello2019challenging}. While these methods demonstrate the value of participant-specific representations and disentangled latent spaces, they generally optimize downstream prediction or personalization rather than explicitly modeling human skill as a persistent, interpretable construct. 

Our work differs by modeling \emph{human skill} as a persistent participant-level representation that is explicitly optimized for construct validity, predictive utility, and interpretable subskill decomposition. By combining participant-specific embeddings, novice--expert basis blending, and counterfactual subskill supervision, \methodname{} learns representations that are stable across repeated observations, predictive across contexts, behaviorally interpretable, and useful for downstream personalization tasks.

\section{Approach}
\label{problem_formulation}
\textbf{Problem Formulation:}
We aim to learn a latent representation of human \emph{skill} from behavioral data.
Let $\mathcal{D}=\{\tau_1,\dots,\tau_N\}$ denote a set of trajectories, where each
$\tau_i=\{x_i^t\}_{t=1}^{T_i}$ is a sequence of feature vectors $x_i^t\in\mathbb{R}^D$
executed in a task context $c\in\mathcal{C}$ (e.g., racetrack or batting condition). We assume contexts $c_i$
  are observed at both train and test time  and that an individual’s skill is  transferable across contexts, while its behavioral expression depends on 
$c$.
Trajectories may include multimodal features such as vehicle telemetry, gaze, or body kinematics.

Our goal is to infer an individual-specific skill embedding $z_s\in\mathbb{R}^d$ that is stable across trajectories and transferable across in-domain contexts (task instances drawn from the same domain).
We distinguish \emph{skill}, a persistent construct, from \emph{performance}~\citep{iso2024theory}, which reflects trial-specific outcomes and is sensitive to situational factors.
We represent skill as \emph{compositional} in which $z_s$ decomposes into interpretable subcomponents $z_s^{(k)}$ corresponding to distinct subskills, consistent with motor learning theories that describe skill as arising from multiple interacting components ~\citep{newell1991motor,anderson1982acquisition}.

To connect  subskills with behavior, we use \emph{skill metrics} $m\in\mathcal{M}$. Skill metrics provide noisy behavioral proxies. These metrics are derived from trajectories, expert annotations, or auxiliary tasks.
Multiple metrics may map to the same subskill and they provide supervision for learning structured representations of $z_s$ (described in Sec \ref{sec:cf}).
Our approach is detailed in Alg.~\ref{alg:sail} and described below.

\begin{figure*}[t]
    \centering
    \includegraphics[width=.8\textwidth]{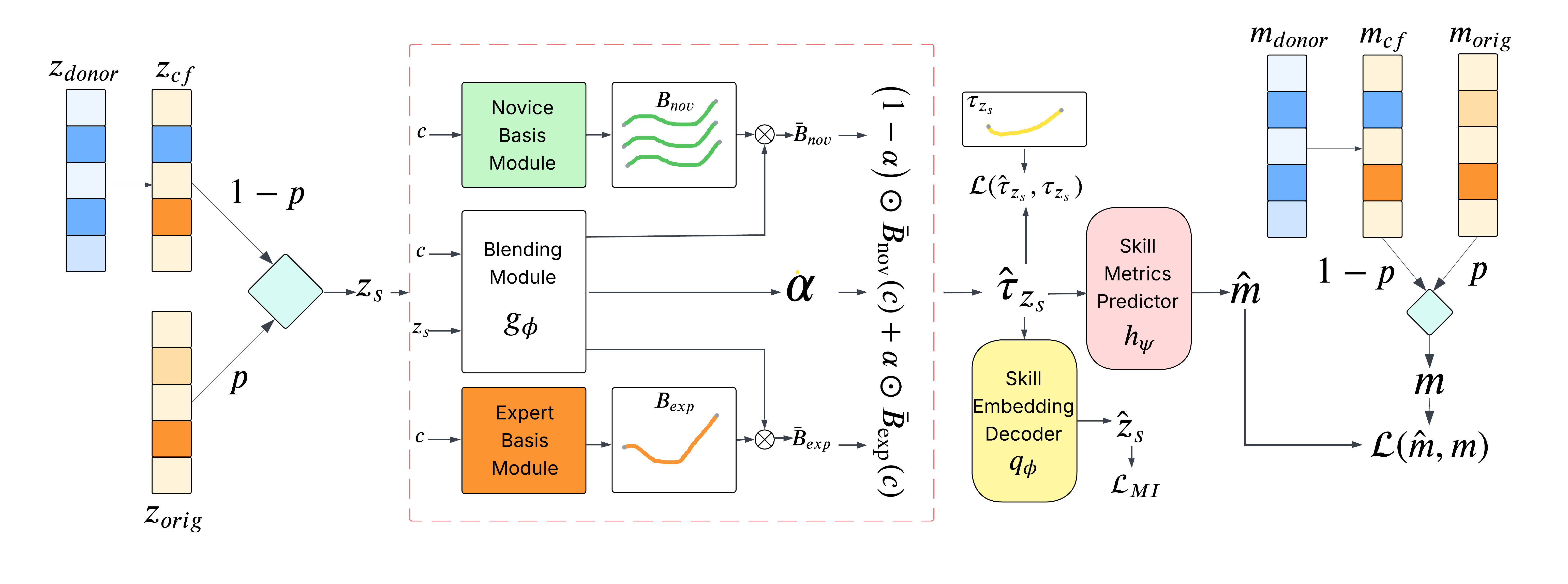}
    \caption{
Overview of \methodname. Each participant is associated with a persistent skill embedding $z_s$ learned across multiple behavioral observations. Rather than decoding trajectories directly, $z_s$ predicts behavior by blending canonical novice and expert basis trajectories, encouraging the representation to capture stable skill-related variation instead of transient behavioral fluctuations. The embedding is partitioned into subskill-specific slices that are supervised using behaviorally grounded skill metrics and disentangled through counterfactual subskill swaps.
}
    \label{fig:overview}
\end{figure*}


\subsection{Participant-Specific Skill Embedding}

Human skill is a persistent characteristic of an individual rather than a single behavioral observation. Inferring skill independently from each trajectory therefore conflates stable ability with trial-specific factors such as fatigue, measurement noise, environmental variation, and strategy. Moreover, trajectory-level embeddings are not inherently tied to the individual who produced them.

To model persistence, \methodname assigns each training participant a learnable skill embedding $z_s\in\mathbb{R}^d$, optimized jointly with the model parameters (Alg.~\ref{alg:sail}, Line~\ref{line:init}). A single embedding is shared across all trajectories from the same participant, pooling evidence across repeated observations to capture stable behavioral tendencies despite trial-to-trial variability. This is conceptually similar to participant embeddings used in recommender systems and speaker recognition~\citep{koren2009matrix,snyder2018x}.

However, persistence alone does not imply skill: a participant embedding could simply summarize average behavior. In the following section, we introduce an additional inductive bias by constraining behavior to be generated relative to canonical novice and expert behavior bases, encouraging $z_s$ to encode expertise rather than arbitrary behavioral variation.

At test time, the model parameters ($g_{\phi}$, $q_{\theta}$, and $h_{\psi}$) are frozen, and only the participant embedding is optimized using the trajectory reconstruction loss.

To prevent collapse of the embedding, we introduce an auxiliary network $q_\theta$ that reconstructs $z_s$ from generated trajectories. This provides a variational lower bound on the mutual information between $z_s$ and predicted behavior and encourages the embedding to encode information that is both behaviorally meaningful and recoverable from observed trajectories~\citep{kingma2013auto,chen2016infogan}.

\color{black}
\begin{algorithm}[tb]\small
\caption{Training \methodname}
\label{alg:sail}
\begin{algorithmic}[1]
\REQUIRE Trajectories $\tau_i$, contexts $c_i$, subskill metrics $m_i$
\STATE Initialize participant embeddings $z_{s,i}\!\sim\!\mathcal{N}(0,0.1)$ \label{line:init}
\FOR{each training iteration}
  \STATE Sample batch of participants and trajectories
  \STATE Predict behavior $\hat{\tau}_{z_s}$ via expert--novice blending (Sec.~3.2)
  \STATE Decode predicted subskill metrics $\hat{m}=h_\psi(\hat{\tau}_{z_s})$
  \STATE Compute total loss $\mathcal{L}=\lambda_{\text{traj}}\mathcal{L}_{\text{traj}}+\lambda_{\text{metric}}\mathcal{L}_{\text{metric}}+\lambda_{\text{MI}}\mathcal{L}_{\text{MI}}$
  \IF{CF step (probability $1-p$)} \label{line:if_cf}
    \STATE Swap $(z_{\text{orig}}^{(k)},m_{\text{orig}}^{(k)})\!\leftarrow\!(z_{\text{donor}}^{(k)},m_{\text{donor}}^{(k)})$ \label{line:swap}
    \STATE Reconstruct CF trajectory $\tilde{\tau}_{\text{orig}}$ from $\tilde{z}_{\text{orig}}$
    \STATE \textbf{Skip trajectory reconstruction loss; apply metric loss only for swapped subskill $k$} \label{line:no_loss}
  \ENDIF
  \STATE Update model parameters and participant embeddings jointly via back-propagation
\ENDFOR
\end{algorithmic}
\end{algorithm}
\color{black}

\subsection{Skill Representation via Novice--Expert Basis Blending}

A participant-specific embedding provides a persistent representation of an individual, but persistence alone does not imply that the embedding represents \emph{skill}. Without additional inductive structure, the embedding may instead encode an individual's average behavior, preferred driving style, or other participant-specific characteristics unrelated to expertise. The challenge is therefore to constrain the representation so that it explains the stable behavioral variation associated with skill while remaining insensitive to transient performance fluctuations.

A straightforward approach is to decode trajectories directly from the learned skill embedding. However, this requires the embedding to account for every aspect of the observed behavior, including variability arising from fatigue, measurement noise, environmental conditions, and idiosyncratic execution. Consequently, the learned representation is encouraged to memorize trajectories rather than isolate the latent factors responsible for expertise.

Instead, we model behavior \emph{relative to canonical novice and expert behaviors}. Rather than asking the embedding to generate a trajectory from scratch, we ask it to explain where an individual's behavior lies relative to representative novice and expert executions for the  task context. This imposes an inductive bias: the embedding need only encode deviations associated with expertise, while the basis trajectories explain common behavioral structure shared across participants. As a result, transient variation and stylistic differences are less likely to be absorbed into the skill representation.

For each task context $c$ (e.g., a racetrack or batting condition), we define sets of canonical novice and expert basis trajectories,
\[
B_{\mathrm{exp}}(c)=\{B_{\mathrm{exp}}^{(i)}(c)\}_{i=1}^{M}, \qquad
B_{\mathrm{nov}}(c)=\{B_{\mathrm{nov}}^{(i)}(c)\}_{i=1}^{K},
\]
where each basis trajectory, $B^{(i)}\in\mathbb{R}^{T\times D}$,  represents a characteristic mode of behavior observed near the extremes of the skill distribution.
The basis trajectories may be obtained from demonstrations, learned jointly with the model, or generated by an optimal controller. In our implementation, we define the expert basis using trajectories from the demonstrator with the strongest domain-specific performance measure and derive the novice basis by applying principal component analysis (PCA) to novice trajectories. The resulting novice bases capture the dominant modes of variation among inexperienced participants, while the expert basis provides a canonical target behavior.

The participant embedding is mapped by $g_\phi$ to expert and novice basis weights, $w_{\text{exp}}(z_s,c)$ and $w_{\text{nov}}(z_s,c)$, together with an interpolation coefficient $\alpha(z_s,c)\in[0,1]^{T\times D}$. The basis weights are constrained to the simplex so that the predicted behavior is expressed as an element-wise convex interpolation of the expert and novice bases (Fig.~\ref{fig:overview}):

\begin{linenomath*}
\begin{equation}\small
\label{eq:expert_novice_blending}
\setlength{\abovedisplayskip}{4pt}
\setlength{\belowdisplayskip}{4pt}
\begin{aligned}
\bar{B}_{\bullet}(z_s,c)
&=
\sum_j
w_{\bullet}^{(j)}(z_s,c)
B_{\bullet}^{(j)}(c),
\quad
\bullet\in\{\mathrm{exp},\mathrm{nov}\},\\
\hat{\tau}_{z_s}
&=
\alpha\odot\bar{B}_{\mathrm{exp}}(z_s,c)
+
(1-\alpha)\odot
\bar{B}_{\mathrm{nov}}(z_s,c).
\end{aligned}
\end{equation}
\end{linenomath*}

Although behavior is expressed as a blend of novice and expert bases, our formulation does not assume that skill lies on a single linear axis. Multiple novice bases capture diverse low-skill strategies (e.g., overcautious, inconsistent, or poorly timed behavior), while multiple expert bases can represent distinct high-skill styles. Furthermore, each subskill independently modulates its own blending coefficients, enabling complex, nonlinear representations of skill.

Unlike direct trajectory decoding, this formulation encourages the embedding to explain behavior in terms of deviations from canonical novice and expert behaviors rather than memorizing every trajectory detail. Behavior prediction serves as a supervision signal that encourages the embedding to capture stable, skill-related structure while remaining predictive of behavior. 

\subsection{Counterfactual Training for Subskill Disentanglement}
\label{sec:cf}

Human skill is inherently \emph{compositional}: coaches reason about performance in terms of multiple interacting subskills and design interventions that target individual deficiencies \citep{ericsson1993role,newell1991motor,wulf2016attentional,anderson1982acquisition}. Accordingly, we partition the latent representation into subskill-specific components that should independently influence the behaviors associated with each subskill. This requires both \emph{disentanglement} (independent latent factors) and \emph{identifiability} (each factor corresponds to a human-recognizable subskill).

Existing disentanglement methods (e.g., InfoGAN, $\beta$-VAE, FactorVAE) encourage statistical independence but do not ensure that latent dimensions correspond to meaningful subskills or support selective behavioral interventions \citep{higgins2017beta,kim2018disentangling,locatello2019challenging}. Conditional supervision associates latent dimensions with labels~\citep{kingma2014semi,sohn2015learning}, but changing a supervised latent need not produce the expected behavioral change.

To address this limitation, we explicitly partition the embedding into subskill-specific slices and train the representation using counterfactual interventions. During training, one subskill slice is replaced with that of another participant while the remaining slices are held fixed. The model is then required to produce behavior that reflects only the substituted subskill, encouraging both disentanglement and identifiability. The embedding space is partitioned as
\[
z_s = \big[\, z_s^{(1)}, \; z_s^{(2)}, \; \dots, \; z_s^{(K)} \,\big],
\]
where each slice $z_s^{(k)} \in \mathbb{R}^{d_k}$ is intended to represent subskill $k$, and $\sum_k d_k = d$.

\color{black}
Reconstructed trajectories $\hat{\tau}_{z_s}$ are passed through a predictor network $h_\psi$ 
to produce subskill metrics $\hat{m}$ (Fig.~\ref{fig:overview}) that serve as behaviorally grounded 
supervision signals during training. Each subskill metric  is  defined in collaboration with domain experts. Skill metrics reflect a measurable behavioral quantity that serves as a proxy for an underlying subskill (e.g., steering smoothness for control
or gaze dispersion for visual attention). These metrics provide weak yet semantically meaningful supervision that anchors each subskill dimension to interpretable aspects of human behavior.

To enforce CF consistency, we perform subskill swaps between a randomly chosen pair of training examples: an \textit{original}     sample (the one being modified) and a \textit{donor} sample 
(the one borrowed from). For a  subskill $k$, we replace the $k$-th 
slice of the original embedding with that of the donor:
\[
\tilde{z}_{\text{orig}}^{(k)} = z_{\text{donor}}^{(k)}, \quad
\tilde{z}_{\text{orig}}^{(\ell)} = z_{\text{orig}}^{(\ell)} \;\; \forall \ell \neq k,
\] and apply the same operation to the associated skill metrics to ensure supervision remains consistent:
\[
\tilde{m}_{\text{orig}}^{(k)} = m_{\text{donor}}^{(k)}, \quad
\tilde{m}_{\text{orig}}^{(\ell)} = m_{\text{orig}}^{(\ell)} \;\; \forall \ell \neq k.
\]
In practice, we interleave CF and standard training. With probability $p$, a batch is trained 
using the regular reconstruction and metric objectives, and with probability $(1-p)$, a batch is trained 
with CF swaps (Alg.~\ref{alg:sail}, Lines~\ref{line:if_cf}--\ref{line:swap}). 
This procedure creates CF examples where the original $z_s$ retains all except one subskill slice  which is borrowed from the donor. Doing so allows the model to learn how isolated subskills should influence predicted behavior and skill metrics.
 \color{black}
 
This approach encourages reconstruction fidelity while also promoting  disentanglement. 
Since no ground-truth trajectory exists for this CF, we do not apply a reconstruction loss to  $\hat{\tau}_{z_s}$ for the swapped  items (\textcolor{black}{Alg. \ref{alg:sail} Line \ref{line:no_loss})}. Instead, the predictor network, $h_\psi$, is required to output the swapped metric for subskill $k$, thus forcing the model to adjust behavior in a way that matches the intervention.

Unlike approaches that impose constraints directly on the latent space \citep{lin2020linear}, our method encourages disentanglement through behavior. By requiring reconstructed trajectories to predict subskill metrics during CF swaps, each latent slice is forced to encode  its designated subskill.
\color{black}
\subsection{Modeling Details and Losses}
The overall training objective encourages predictive accuracy, semantic alignment, and disentanglement:
\begin{equation}
\label{eq:total_loss}
\mathcal{L}
= \lambda_{\text{traj}} \mathcal{L}_{\text{traj}}
+ \lambda_{\text{metric}} \mathcal{L}_{\text{metric}}
+ \lambda_{\text{MI}} \mathcal{L}_{\text{MI}} .
\end{equation}

Here, $\mathcal{L}_{\text{traj}}(\hat{\tau}_{z_s}, \tau)$ is a trajectory reconstruction loss between the predicted trajectory $\hat{\tau}_{z_s}$ and the observed trajectory $\tau$.
$\mathcal{L}_{\text{metric}}(h_\psi(\hat{\tau}_{z_s}), m)$ supervises behaviorally grounded subskill metrics by comparing predicted metrics $h_\psi(\hat{\tau}_{z_s})$ to targets $m$.
Finally, $\mathcal{L}_{\mathrm{MI}}
=
-\mathbb{E}_{\hat{\tau}\sim p_\phi(\hat{\tau}\mid z_s,c)}
\left[\log q_\theta(z_s\mid\hat{\tau})\right]$ is a mutual-information objective. The mutual-information term is implemented by re-encoding the predicted trajectory $\hat{\tau}_{z_s}$ through $q_\theta$ to obtain $\hat{z}_s$, and encourages the embedding $z_s$ to be recoverable from generated behavior.
During CF training steps, $\mathcal{L}_{\text{traj}}$ is omitted since no ground-truth trajectory exists for the swapped embedding, and only the metric loss for the swapped subskill is applied.

Our model integrates  skill embeddings with trajectory and context encoders from established sequence architectures. 
The trajectories are predicted via two decoders, which produce elementwise blending weights for basis blending. $h_{\psi}$ uses an LSTM to predict the skill metrics from $\hat{\tau}_{z_s}$. 
\color{black}

\section{Domains and Datasets}
\label{sec:domains}
We evaluate \methodname in two domains with substantially different movement dynamics and subskill structure: high-performance racing and baseball batting. Both domains require coordinated mastery of multiple interacting subskills and provide measurable behavioral outcomes, making them suitable testbeds for evaluating human skill representations.\footnote{The human-subjects data collection protocol was approved by WCG IRB in June 2023.}
\subsection{High-Performance Racing}
High-performance racing is a compelling domain for studying skill because it requires the integration of multiple subskills to achieve  mastery. We focus on six core subskills identified by expert coaches and prior work \citep{Schrum2025}: (i) vehicle handling, (ii) gaze control, (iii) know-how, (iv) control inputs, (v) physical ability, and (vi) perceptual ability.  \textcolor{black}{These subskills correspond  to how professional coaches diagnose driver weaknesses and training interventions.}

Each trajectory $\tau_i$ consists of vehicle pose, speed, and control signals downsampled to 100 points per track segment. The context $c$ for this dataset refers to the racetrack that the trajectory was performed on. We collected a dataset of racing trajectories from 95 participants spanning novices to experts, using a  driving simulator. We collected data in two phases: 70 participants each completed at least ten laps on a single track modeled after a nearby raceway, and 25 participants completed four laps on each of four distinct tracks at the same venue. This design provided both breadth (a large participant pool) and depth (multiple laps and multiple contexts). In total we collected 1545 laps. 

To connect observed behavior to underlying subskills, we used a set of behaviorally grounded skill metrics $m \in \mathcal{M}$,  defined in collaboration with expert coaches in prior work \citep{Schrum2025}. Each metric is derived from a task designed to probe a specific subskill. For example, peak lateral g-force in a skidpad drill reflects vehicle handling, 
gaze fixation  during driving sessions reflects gaze policy, 
and written test scores reflect  know-how of racing lines and other HPD techniques. These metrics (among others) provide partial, noisy evidence about latent subskills and provide the supervision signals necessary for learning disentangled representations of $z_s$.



\subsection{Baseball Hitting}
We applied \methodname to a supplemental dataset of baseball hitting collected from 13 players on a competitive adult team in a semi-professional league. While all participants were experienced players, they were not at the level of an expert benchmark and thus exhibited 
substantial variation across subskills. In collaboration with a coach, one highly skilled participant was identified as an expert and used to define the canonical expert basis for blending, while the remaining players provided a diverse set of  trajectories. In total, 74 batting trials were recorded, across both pitching machine sessions and tee batting conditions. 
Whole-body kinematics of swing motions were captured using an optical motion capture system. The coach identified three core subskills and associated metrics of hitting: (i) the \emph{kinematic chain}, or the sequential transfer of momentum across body segments; (ii) \emph{pelvis pausing}, or the ability to momentarily stabilize the pelvis to build rotational 
power; and (iii) \emph{thigh pausing}, or the controlled deceleration of the lead thigh. The contexts $c$ are tee batting and machine-pitch batting.

To address the limited size of the dataset, we generated synthetic participants by applying trajectory augmentations (time warping, noise injection, and scaling) to the data. For players with both tee and machine-pitch trials, we estimated a global offset between conditions and used it to synthesize additional regular swings. This produced artificial batting trials that preserved the underlying structure  while introducing  diversity. 

To our knowledge, there are no existing datasets that capture multimodal behavioral signals and skill metrics that are comparable in richness to our racing dataset. Unlike the racing dataset,  the baseball dataset is smaller, narrower in subskill coverage, and augmented with synthetic trials. We therefore treat this baseball dataset as a supplemental, secondary domain to test the generality of \methodname.

\section{Results}

\begin{table*}[t]
\caption{Results  in Racing (R) and Baseball (B). 
Higher is better for \up, lower is better for \down. Bold = best. Values report mean (standard error) across evaluation folds. }
\label{tab:leaderboard-two-domains}
\centering
\small
\resizebox{\textwidth}{!}{
\begin{tabular}{l *{14}{c}}
\toprule
& \multicolumn{2}{c}{\textbf{\methodname (ours)}} 
& \multicolumn{2}{c}{\textbf{\methodname w/o CF}} 
& \multicolumn{2}{c}{\textbf{\methodname w/o basis}} 
& \multicolumn{2}{c}{\textbf{SimCLR}} 
& \multicolumn{2}{c}{\textbf{$\beta$-VAE}} 
& \multicolumn{2}{c}{\textbf{AE}} 
& \multicolumn{2}{c}{\textbf{AE-LC}} \\
\cmidrule(lr){2-3} \cmidrule(lr){4-5} \cmidrule(lr){6-7} 
\cmidrule(lr){8-9} \cmidrule(lr){10-11} \cmidrule(lr){12-13} \cmidrule(lr){14-15}
& \multicolumn{1}{c}{R} & \multicolumn{1}{c}{B}
& \multicolumn{1}{c}{R} & \multicolumn{1}{c}{B}
& \multicolumn{1}{c}{R} & \multicolumn{1}{c}{B}
& \multicolumn{1}{c}{R} & \multicolumn{1}{c}{B}
& \multicolumn{1}{c}{R} & \multicolumn{1}{c}{B}
& \multicolumn{1}{c}{R} & \multicolumn{1}{c}{B}
& \multicolumn{1}{c}{R} & \multicolumn{1}{c}{B} \\
\midrule
\multicolumn{15}{l}{\emph{Construct Validity}} \\
Silhouette (\up)                 & .72 (.08) & & \textbf{.77 (.16)}  &  & .67 (.07) &  & .40 (.44) &  & .74 (.16) &  & .75 (.18) &  & .67 (.18) & \\
Test--retest similarity (\up)
& \textbf{.995 (.003)} & \textbf{1.000 (.003)}
& .995 (.001) & .998 (.001)
& .990 (.004) & .995 (.003)
& .928 (.10) & .958 (.119)
& .928 (.02) & .498 (.330)
& .839 (.12) & .979 (.090)
& .891 (.22) & .998 (.001) \\

\textbf{Overall Construct Score} (\up)
& 1.86 & \textbf{1.00}
& \textbf{2.0} & .994
& 1.69 & .992
& .57 & .919
& 1.49 & 0.00
& .95 & .961
& 1.06 & .996 \\

\midrule
\multicolumn{15}{l}{\emph{Predictive Utility}} \\

Behavior prediction (RMSE \down)
& \textbf{2.75 (.12)} & .161 (.070)
& 2.75 (.12) & \textbf{.155 (.075)}
& 5.05 (.26) & .176 (.088)
& 4.15 (.99) & .204 (.087)
& 4.48 (.36) & .479 (.158)
& 4.61 (.38) & .257 (.117)
& 4.87 (.55) & .191 (.093) \\

OOC generalization (RMSE \down)
& \textbf{6.37 (3.0)} & .120 (.053)
& 6.50 (3.1) & \textbf{.115 (.061)}
& 12.77 (3.1) & .115 (.059)
& 10.27 (5.0) & .395 (.090)
& 12.51 (3.5) & .768 (.128)
& 12.42 (3.1) & .152 (.087)
& 14.12 (2.8) & .153 (.094) \\

\textbf{Overall Predictive Score} (\up)
& \textbf{2.0} & 1.94
& 1.98 & \textbf{2.00}
& .17 & 1.89
& .89 & .98
& .46 & 0.00
& .41 & .81
& .08 & .95 \\

\midrule
\multicolumn{15}{l}{\emph{Disentanglement \& Interpretability}} \\

Alignment Ratio (AR \up)
& \textbf{3.25 (.79)} & \textbf{0.758 (.025)}
& 1.24 (.083) & 0.416 (.023)
& 1.11 (.21) & 0.301 (.057)
& 1.73 (.78) & 0.087 (.074)
& 1.12 (.11) & 0.046 (.052)
& 1.05 (.14) & 0.387 (.030)
& 2.40 (.17) & 0.724 (.029) \\

Targeted Change Index (TCI \up)
& \textbf{.93 (.06)} & \textbf{0.146 (.019)}
& .89 (.10) & 0.139 (.013)
& .73 (.05) & 0.128 (.010)
& .45 (.10) & 0.132 (.011)
& .63 (.03) & 0.140 (.021)
& .78 (.06) & 0.134 (.022)
& .63 (.03) & 0.144 (.023) \\

Relative Influence Ratio (RIR \up)
& \textbf{2.11 (.46)} & \textbf{1.244 (.054)}
& 1.76 (.19) & 1.129 (.177)
& 1.67 (.22) & 1.104 (.107)
& 1.85 (1.14) & 1.101 (.076)
& 1.86 (.23) & 0.731 (.267)
& 1.61 (.32) & 1.203 (.049)
& 1.56 (.28) & 1.122 (.072) \\

\textbf{Overall Interpretability Score} (\up)
& \textbf{3.0} & \textbf{2.85}
& 1.37 & 1.82
& .81 & 1.08
& .83 & .97
& .95 & 1.00
& .78 & 1.68
& 1.20 & 2.46 \\
\end{tabular}%
}
\end{table*}

\begin{figure}
    \centering
    \includegraphics[width=0.42\textwidth]{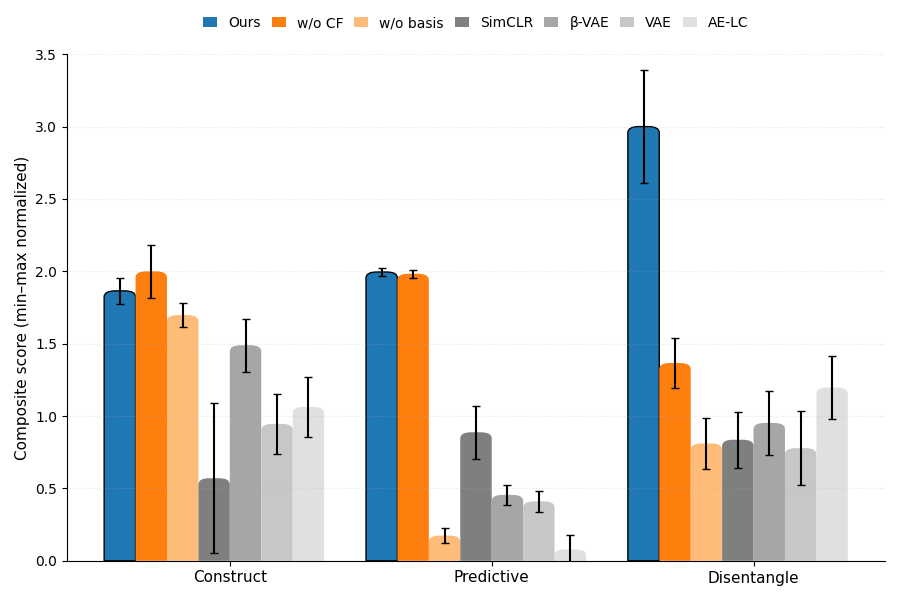}
    \caption{\textcolor{black}{Composite scores across the three desiderata in Racing. Bars show performance of  (\methodname), 
    ablations, and baselines. 
    Higher is better for all desiderata.}}
    \label{fig:composite-barplot}
\end{figure}

 Since our contribution is a representation learning method rather than a policy-learning or imitation-learning algorithm, we compare \methodname against established representation learning baselines designed to evaluate latent representations. Composite scores are shown for Racing (Fig.~\ref{fig:composite-barplot}), our primary domain, while Baseball results are reported in Table~\ref{tab:leaderboard-two-domains} as a supplemental domain.

\textbf{SimCLR (contrastive baseline).} A self-supervised method that uses contrastive losses to encourage invariance within an individual. We adapt SimCLR to trajectory data to test whether a  contrastive objective is sufficient for extracting skill-relevant embeddings~\citep{chen2020simclr}.

\textbf{$\beta$-VAE (disentanglement baseline).} An extension of the VAE with stronger KL regularization that encourages factorized latents. We include $\beta$-VAE as a  disentanglement method to test whether standard disentanglement approaches yield interpretable subskills~\citep{higgins2017beta}.

\textbf{AE (autoencoder baseline).} A standard trajectory
autoencoder~\citep{hinton2006reducing} that captures per-trial
variability but is not designed to model persistent skill or
subskill structure.

\textbf{AE-LC (AE with linear constraints).} An extension of the AE framework that incorporates linear constraints derived from subskill metrics to encourage semantically meaningful and identifiable latents. We include this method to test whether metric-based structure alone can recover interpretable subskills compared to \methodname~\citep{lin2020linear}.

\textbf{Ablation: without CF training (\methodname w/o CF).} This ablation removes the CF swap objective and trains only with behavioral prediction via expert–novice basis blending to isolate the contribution of CF supervision.

\textbf{Ablation: without expert–novice basis and CF training (\methodname w/o basis).} This ablation decodes trajectories directly from the skill embedding without basis blending or CF supervision to test whether the basis decomposition is necessary for isolating skill-related variation from  transient factors.

For all baselines that operate at the trial level (SimCLR, AE, $\beta$-VAE, AE-LC), we extract embeddings per trajectory and  pool  across laps for each participant which produces a participant-level embedding comparable to our method.
We evaluate our approach and baselines along the three desiderata introduced in Section~\ref{sec:intro}:  
(1) construct validity, (2) predictive utility , and (3) disentanglement and interpretability. For each desideratum, we compute a composite score by min–max normalizing each metric across methods, reversing lower-is-better metrics, and summing the normalized values. We also evaluate whether the learned representation improves a downstream AI coaching model and validate the representation against a professional coach's ratings.

\subsection{Construct Validity}
\label{sec:results-construct}
We evaluate \emph{construct validity} by measuring whether the learned embedding captures stable, skill-relevant structure rather than transient fluctuations  (Table~\ref{tab:leaderboard-two-domains}). 
We operationalize construct validity as stable within-participant and discriminative across-skill representations. Because no coaching occurred during data collection, we assume participants' underlying skill remained approximately constant. We evaluate construct validity using silhouette score and test–retest similarity.

\begin{itemize}
    \item \textbf{Silhouette score} $(\uparrow)$: clustering quality by skill group.
    \item \textbf{Test--retest similarity} $(\uparrow)$: stability of embeddings across repeated trials.
\end{itemize}

\textbf{Discussion:} As shown in Figure~\ref{fig:composite-barplot} and Table~\ref{tab:leaderboard-two-domains}, among the evaluated methods, \methodname{}  achieves strong overall construct validity across both racing and baseball. \methodname{} produces highly stable embeddings (test--retest similarity of 0.995 in racing and 1.000 in baseball), indicating that $z_s$ captures persistent aspects of skill rather than trial-level variability. The no-CF ablation performs similarly, suggesting that counterfactual supervision preserves construct validity while primarily benefiting interpretability. In contrast, removing the novice--expert basis reduces clustering quality, likely because the embedding captures more trial-specific variation. Overall, these results suggest that participant-level embeddings and basis blending contribute to learning stable skill representations. We do not report silhouette scores for baseball because discrete skill labels are unavailable.

\subsection{Predictive Utility}
\label{sec:results-predictive}
We next investigate \emph{predictive utility} by evaluating if the learned skill embeddings support accurate trajectory prediction  within and across contexts. Predictive utility is a key desideratum, because it indicates whether \methodname can be used to anticipate behavior for a given skill and how behavior will change under novel conditions. 
We evaluate predictive utility using  in-context (trained and tested on the same set of racetracks, with held-out trials) and out-of-context (trained on one racetrack, tested on different track) prediction metrics.
\begin{itemize}
    \item \textbf{In-context prediction (RMSE $\downarrow$):} trajectory accuracy within the same context.  
    \item \textbf{Out-of-context prediction (RMSE $\downarrow$):} generalization to novel contexts.  
\end{itemize}

\textbf{Discussion:} As shown in Figure~\ref{fig:composite-barplot} and Table~\ref{tab:leaderboard-two-domains}, \methodname{} achieves the best  predictive performance in racing and remains competitive in baseball. Removing counterfactual (CF) supervision has little effect on prediction, indicating that CF primarily improves interpretability. In contrast, removing novice--expert basis blending nearly doubles prediction error in racing, suggesting that the basis is the primary source of predictive generalization by separating stable skill from transient  variation. Together, these results indicate that participant-level embeddings and basis blending drive predictive utility, while CF selectively improves disentanglement.

\subsection{Disentanglement and Interpretability}
\label{sec:results-disentangle}

Finally, we evaluate whether the representation decomposes into interpretable subcomponents that correspond to distinct subskills via alignment ratio, targeted change index, and relative influence ratio metrics (Table~\ref{tab:leaderboard-two-domains}): Together, these metrics evaluate disentanglement along three complementary axes: semantic alignment (AR), selective intervention effects (TCI), and relative influence on  outputs (RIR).

\color{black}
\begin{itemize}
    \item \textbf{Alignment Ratio (AR $\uparrow$):} measures how well each subskill slice $z_s^{(k)}$ predicts its intended metrics compared to non-target ones, indicating subskill–metric correspondence \citep{eastwood2018framework}. 
    \item \textbf{Targeted Change Index (TCI $\uparrow$):} operationalizes  the idea of intervention selectivity described in \citet{bengio2020causal} and \citet{scholkopf2021toward} and quantifies the effect of CF swaps by checking whether trajectory changes are concentrated in the targeted features, with higher values reflecting more selective control.  
    \item \textbf{Relative Influence Ratio (RIR $\uparrow$):} complements TCI by perturbing one subskill at a time and measuring how much this changes an expected behavioral feature, relative to the change induced by perturbing other subskills.
\end{itemize}

\color{black}
\textbf{Discussion:} Figure~\ref{fig:composite-barplot} and Table~\ref{tab:leaderboard-two-domains} show that \methodname{} consistently achieves the strongest disentanglement across both domains. Removing counterfactual (CF) supervision substantially reduces all interpretability metrics, demonstrating that CF training is the primary mechanism for learning semantically meaningful subskill representations. Although AE-LC incorporates explicit metric supervision, it consistently underperforms \methodname{}, indicating that supervision alone is insufficient to produce behaviorally grounded, selectively controllable representations.

\begin{table}[t]
\centering
\small
\caption{Downstream coaching-instruction prediction (mean $\pm$ SE
over 15 participant-held-out folds). \methodname{} significantly
outperforms the trial-time baseline on weighted F1 (paired
$t(14){=}2.50$, $p{=}.025$) and accuracy ($p{=}.028$); on macro F1,
\methodname{} is the only condition that significantly improves over
no conditioning ($p{<}.001$).}
\label{tab:coaching}
\begin{tabular}{lccc}
\toprule
Conditioning & Weighted F1 \up & Macro F1 \up & Acc.\ \up \\
\midrule
None       & .541 (.015) & .504 (.018) & .538 (.014) \\
Trial time & .573 (.014) & .520 (.020) & .567 (.014) \\
\methodname{} (ours) & \textbf{.595 (.014)} & \textbf{.537 (.018)} & \textbf{.589 (.013)} \\
\bottomrule
\end{tabular}
\end{table}
\subsection{Skill-Informed Coaching}
Finally, we evaluate the downstream utility of \methodname{} by incorporating the learned skill representation into a previously proposed imitation-learning model for predicting instructor feedback \citep{Anonymous2025Coach}. We augment the original model with a frozen participant embedding computed from the participant's previous four laps, and compare against conditioning on a scalar baseline (trial time) over the same window.
We evaluate on two previously collected simulator coaching datasets comprising 38 participants \citep{Anonymous2025Dataset, Anonymous2025Skill}. As shown in Table~\ref{tab:coaching}, conditioning on the \methodname{} embedding yields a 10.0\% relative improvement in weighted F1 over the unconditioned model and significantly outperforms trial-time conditioning on weighted F1 and accuracy. Unlike trial time, \methodname{} also significantly improves macro F1, suggesting that the representation captures information beyond overall ability and improves prediction across both common and infrequent instruction categories. These results suggest that access to $z_s$ enables more accurate prediction of both when and what instructors will coach.

\subsection{External Validation by Professional Coach}

To assess whether the learned representation aligns with expert human
judgment, we compare \methodname{} skill estimates against
independent ratings from a professional driving coach who evaluated
22 held-out participants over the course of a coaching study. These
ratings were collected independently of the drill-based metrics used
for training supervision. To obtain a scalar overall skill estimate from \methodname{}, we project each
participant's embedding onto the direction between the mean novice
and expert embeddings and convert the resulting position to a
percentile, where 0 and 100 correspond to the novice and expert
reference points, respectively. The coach's overall skill ratings agree
strongly with \methodname{}'s overall skill estimate (Spearman
$\rho = 0.81$, $p < .001$, 95\% bootstrap CI $[0.56, 0.94]$, $n=22$),
indicating that the embedding aligns well with human expert judgments of skill.  Because these coach ratings and downstream coaching labels were not used as supervision for learning the representation, these results provide evidence that SAIL captures information beyond the behaviorally grounded metrics used during training.

\section{Limitations}
Our evaluation is limited by dataset scale and scope, particularly in the baseball domain where data are small and augmented.
The method also depends on noisy, predefined subskill metrics, and assumes a smooth novice–expert continuum that may miss certain strategies.   While  metrics are defined in collaboration with domain experts, they may be incomplete or biased.  Our evaluation focuses on practical properties of a useful skill representation—stability, predictive utility, interpretability, and agreement with expert judgment—rather than establishing a unique or complete computational definition of human skill. Future work should investigate additional forms of construct validation and longitudinal studies of skill acquisition. 
\bibliography{aaai2027}


\end{document}